%% file: root.tex
\documentclass[sigconf]{acmart}
\usepackage{todonotes}
\usepackage[nolist]{acronym}
\usepackage{cleveref}
\usepackage{placeins}

\AtBeginDocument{%
  }

\begin{document}
\input{acro}

\title{Navigating Dimensionality through State Machines in Automotive System Validation}

\author{Laurenz Adolph}
\email{adolph@fzi.de}
\affiliation{%
  \institution{FZI Research Center for Information Technology}
  \city{Karlsruhe}
  \state{Baden-W{\"u}rttemberg}
  \country{Germany}
}

\author{Barbara Schütt}
\email{schuett@fzi.de}
\affiliation{%
  \institution{FZI Research Center for Information Technology}
  \city{Karlsruhe}
  \state{Baden-W{\"u}rttemberg}
  \country{Germany}
}

\author{David Kraus}
\email{d.kraus@kit.edu}
\affiliation{%
  \institution{Karlsruhe Institute of Technology}
  \city{Karlsruhe}
  \state{Baden-W{\"u}rttemberg}
  \country{Germany}
}

\author{Eric Sax}
\email{eric.sax@kit.edu}
\affiliation{%
  \institution{Karlsruhe Institute of Technology}
  \city{Karlsruhe}
  \state{Baden-W{\"u}rttemberg}
  \country{Germany}
}

\renewcommand{\shortauthors}{Adolph et al.}

\begin{abstract}

The increasing automation of vehicles is resulting in the integration of more extensive in-vehicle sensor systems, electronic control units, and software. Additionally, vehicle-to-everything communication is seen as an opportunity to extend automated driving capabilities through information from a source outside the ego vehicle. However, the validation and verification of automated driving functions already pose a challenge due to the number of possible scenarios that can occur for a driving function, which makes it difficult to achieve comprehensive test coverage. Currently, the establishment of \ac{SOTIF} mandates the implementation of scenario-based testing. The introduction of additional external systems through vehicle-to-everything further complicates the problem and increases the scenario space. In this paper, a methodology based on state charts is proposed for modeling the interaction with external systems, which may remain as black boxes. This approach leverages the testability and coverage analysis inherent in state charts by combining them with scenario-based testing. The overall objective is to reduce the space of scenarios necessary for testing a networked driving function and to streamline validation and verification. The utilization of this approach is demonstrated using a simulated signalized intersection with a roadside unit that detects vulnerable road users.

\end{abstract}





\maketitle

\input{sections/01_Introduction}
\input{sections/02_State_of_the_art}

\input{sections/03_Concept}

\input{sections/04_Implementation}
\input{sections/05_Evaluation}

\input{sections/06_Conclusion}

\FloatBarrier
\bibliographystyle{ACM-Reference-Format}
\bibliography{statecharts}

\include{sections/07_Appendix}

\end{document}

%% file: acro.tex
\begin{acronym}
    \acro{ECU}{Electronic Control Unit} 
    \acro{OEM}{Original Equipment Manufacturer}
    \acro{ISO}{International Organization for Standardization}
    \acro{ASIL}{Automotive Safety Integrity Level} 
    \acro{SAE}{Society of Automotive Engineers}
    \acro{AD}{Autonomous Driving}
    \acro{ADS}{Autonomous Driving Systems}
    \acro{ADAS}{Advanced Driver Assistance Systems}
    \acro{OTA}{Over The Air}
    \acro{SW}{software}
    \acro{V2X}{Vehicle-to-Everything}
    \acro{V2I}{Vehicle-to-Infrastructure}
    \acro{V2V}{Vehicle-to-Vehicle}
    \acro{V2P}{Vehicle-to-Pedestrian}
    \acro{V2C}{Vehicle-to-Cloud}
    \acrodef{UML}[UML]{Unified modelling language}
    \acrodef{E/E}[E/E-architecture]{Electrical/Electronic architecture}
    \acrodef{HW}[HW]{Hardware}
    \acro{HPC}{High-Performance-Computing}
    \acro{TRL}{Technology Readiness Level}
    \acro{VV}[V\&V]{Validation and Verification}
    \acro{ACC}{Adaptive Cruise Control}
    \acro{RSU}{Roadside Unit}
    \acro{VRU}{Vulnerable Road User}
    \acro{CASE}{Connected, Autonomous, Shared, Electric}
    \acro{SOTIF}{Safety Of The Intended Functionality}
    \acro{UML}{Unified Modeling Language}
    \acro{ODD}{Operational Design Domain}
    \acro{ITS}{Intelligent Transportation Systems}
    \acro{VANET}{Vehicular Ad Hoc Network}
    \acro{CCP}{Coupon Collector's Problem}
    \acro{DDT}{Dynamic Driving Task}
    \acro{C-V2X}{Cellular V2X}
    \acro{DSRC}{Dedicated Short Range Communication}
    \acro{CAM}{Cooperative Awareness Message}
    \acro{ETSI}{European Telecommunications Standards Institute}
    \acro{BSM}{Basic Safety Message}
    \acro{MEC}{Mobile Edge Computing}
    \acro{SWC}{Software Component}
    \acro{TARA}{Threat Analysis and Risk Assessment}
    \acro{SOP}{Start of Production}
    
\end{acronym}

%% file: sections/01_Introduction.tex
\section{Introduction} \label{sec:introduction}
In the fast-changing field of automated and autonomous driving, the development of \ac{ADAS} and \ac{AD} capabilities is being driven by the emergence of connected systems that use \ac{V2X} communication and edge applications. 
The implementation of higher levels of automation necessitates the secure detection of other traffic participants. 
Additional information from sensors external to the vehicle expands the field of view, thereby aiding in the detection of objects within various situations. 
The challenging verification and validation of these cyber-physical \ac{ADAS} and \ac{AD} systems including sensors is a significant bottleneck in achieving Level 4 automation. 
Degrees of freedom of Level 4 automation make it impractical to comprehensively model and test the entire driving function across all possible situations and systems.
Scenario-based approaches have been introduced to address this challenge, exemplified by concepts such as \ac{SOTIF}. 
Despite the usefulness of scenario-based testing, the number of potential scenarios the driving function could encounter remains an unsolved challenge, as exhaustive testing on the road has been shown to be implausible \citep{maurer_freigabe_2015}. \citet{kaur_towards_2020} use formal methods for the validation of \ac{V2X} systems. 
These, however, require specialized knowledge about all involved systems and are not easily applicable to the predominant, scenario-based approach to \ac{VV}. 

\begin{figure}
    \centering
    \includegraphics[width=0.9\linewidth]{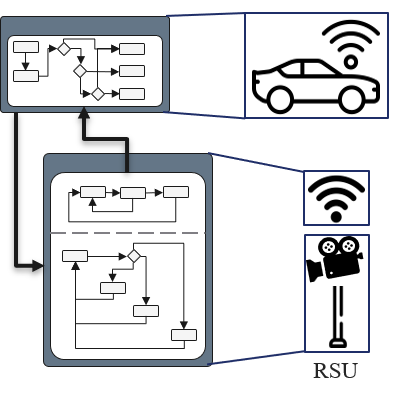}
    \caption{Connected systems as state charts}
    \label{fig:teaser}
\end{figure}

Breaking down the cyber-physical system into smaller, more manageable subsystems allows for more effective testing. This approach requires comprehensive modeling of the interactions of connected sensor systems, even when only partial knowledge of third-party systems like \acp{RSU} is available. 
This work proposes modeling the interaction of the driving function with other systems in a connected environment using state machines. 
The methodology aims to provide a more efficient and manageable path for \ac{VV} by focusing on the states and interactions of individual subsystems rather than attempting to model the entire function. A combined approach, integrating state charts with scenario-based methods, provides a solution to reconcile testability with dimensionality. 
State charts can facilitate the selection, search, and definition of specific test cases for the cyber-physical system.
The methodology will be examined using a simulated intersection with a modeled \ac{RSU} locating other traffic participants and a traffic light broadcasting its current state as shown in \cref{fig:teaser}.

\textit{Contribution:} The paper aims to contribute to the ongoing discourse on the advancement of \ac{VV} and testing methodologies in the context of connected automotive technologies by: 
\begin{itemize}
    \item Modeling interactions in connected driving functions using state charts
    \item Utilization of state chart combinatorics for test coverage analysis
    \item Leveraging the testability of state charts to derive test cases for the \ac{VV} of highly automated driving
\end{itemize}

%% file: sections/02_State_of_the_art.tex
\section{State of the Art}\label{sec:state}

\subsection{Scenario-based testing and \ac{SOTIF}} \label{subsec:scenario-based}

To ensure that the function under development meets the \ac{DDT} in the specified \ac{ODD} for a high level of automation, a structured and thoughtful process for \ac{VV} is required. In Europe, the most common approach to \ac{VV} is \ac{SOTIF}, which is described in general through ISO 26262 \citep{iso_26262} and in an automotive-specific context through ISO/PAS 21448 \citep{technical_committee_isotc_22_road_vehicles_subcommittee_sc_32_electrical_and_electronic_components_and_general_system_aspects_isopas_2022}. It aims to describe the interaction between a vehicle and other participants through scenarios. These scenarios may include multiple snapshots of dynamic events, information about the environment, or other relevant details. 
The PEGASUS-Project \cite{pegasus_consortium_and_others_pegasus_nodate} proposes a standardized method of describing scenarios. The scenario-based method designs and organizes potential situations the vehicle may encounter during operation into scenarios. This condensed set of scenarios can then be tested regarding the \ac{DDT}. 
To determine if a vehicle can meet the requirements of the \ac{DDT} in a particular \ac{ODD}, one alternative is to field-test the automated driving function. This, however, requires an already existing driving function and 2-11 billion kilometers of real-world testing, as proposed by \cite{maurer_freigabe_2015, kalra_driving_2016}.

Although there are several different definitions for the terms scene and scenario, this paper follows the definitions of \citet{ulbrich_defining_2015}.
Their definition states that a scene is delineated in terms of all static and dynamic elements and the associated properties at a specific point in time. 
Consequently, a scenario is defined as a temporal sequence of successive scenes.
As stated by \citet[pp.~83]{wood_safety_2019}, scenario-based testing represents a suitable approach to supplementing the distance-based approach of real-world driving, thus reducing the necessary mileage.
The scenario-based approach encompasses a series of techniques and strategies employed during the testing process, which collectively facilitate the acquisition of information and the formulation of statements regarding the quality of the system under test.

The identification of novel scenarios is a mandatory step in the process of defining new test cases designed to assess the safety of an automated driving system, and is suggested by several standards and regulations \citep{technical_committee_isotc_22_road_vehicles_subcommittee_sc_32_electrical_and_electronic_components_and_general_system_aspects_isopas_2022, council_of_european_union_regulation_2018, council_of_european_union_commission_2022}. 
There are many ways to generate, extract, or find new or critical scenarios from recorded data or from already existing scenarios \citep{schutt20231001}.

Scenario coverage addresses the question whether concrete scenarios are suitable for representing logical scenarios, and it is a crucial component when exploring scenarios.
One approach to measure scenario coverage was proposed by OpenSCENARIO DSL \citep{asam_openscenario_asam_2020} and Foretellix \citep{foretellix_measurable_2020}, is related to the grid search and defines concrete scenarios stepwise for a given parameter range as in an approach by \citet{mori2022inadequacy}.
In this work, the authors state, grids with defined step sizes may overlook critical scenarios even when approaching a relatively narrow coverage.

\citet{hauer_did_2019} proposed an alternative approach that compares the problem of \textit{Have all scenario types been tested?} with the \ac{CCP}, a model from probability theory and related to the urn problem.
The idea behind this is that there is a given set of $N$ coupons, baseball cards, or similar items, and a collector has to draw each type of card with a constant probability.
There are two variations of the classical \ac{CCP}: (1) the probability differs for some types (e.g., McDonald's Monopoly); or (2) the number of types $N$ is not known a priori.
Unfortunately, no analytical solution for the \ac{CCP} exists \citep{hauer_did_2019}, and Monte Carlo approaches are used.

\subsection{State chart} \label{subses:statchart}
State charts can be used to model objects and their possible states and transitions. 
They are formalized using the \ac{UML} \cite{kecher_uml_2007}. 
State charts provide the ability to model conditional transitions, orthogonal states, and the behavior of the state machine at runtime for modeling the behavior of networked systems. 
In contrast to Markov and Hidden Markov models, where transitions between states occur with a given probability and the transition depends only on the current state of the model, the transitions of state charts are event-based and can contain guards that act as conditional transitions. 
This allows interactions between subsystems to be modeled without the need to model each subsystem in its entirety. 
State machines are easy to test because of their finite states and deterministic behavior and can, therefore, be a way of reducing the effort required to \ac{VV} interconnected systems.

\subsection{Current state \ac{V2X}}\label{subsec:v2x} 

In order to enable seamless V2X communication, vehicles, \ac{RSU} as well as other devices that need to interact within the V2X domain, have to agree on a well-defined communication standard.
Within the range of possibilities, 3GPP \ac{C-V2X} - a cellular based application based on 4G/5G \cite{3GPP} and WLAN IEEE 802.11p \cite{802.11p} (also \ac{DSRC}), which uses wireless transmission via 5.9 GHz - are the two main emerging technologies used for \ac{V2X} communication.
To increase widespread adaption, ventures from the EU, the USA, and China are being made to define the contents of \ac{V2X} data transmission.
As for the EU, the \ac{ETSI} has made efforts defining a European communication standard, which is based on either IEEE 802.11p or \ac{C-V2X}. \ac{ETSI} also defined a \ac{V2X} application called \ac{CAM}, which includes message contents and sending trigger conditions \cite{ETSI.2023.CAM}.
However, the USA and China are adapting the \ac{BSM} message \cite{SAE.BSM} introduced by \ac{SAE}.
Through the versatility of \ac{V2X}-networks, security becomes a major concern. With the integration of external devices, new, unforeseen attack patterns can emerge.
In order to protect the intended \ac{AD} functionalities as well as ensuring data-privacy, efficient measures after \ac{SOP} have to be established.
This can be achieved through the extension of \ac{TARA} as shown in \cite{GAP.TARA}, where a more efficient incident handling is proposed. 
Through a triage process, cybersecurity incidents are prioritized based on the risk assessment of the attack incident.

\subsection{Edge Computing and outsourcing of functions}\label{subsec:edge}
Transferring \acp{SWC} into nearby devices through \ac{MEC} can reduce energy consumption, as the computational load within the vehicle is reduced. To enable this, task allocating processes have to be developed.
One approach is the Mutlihop Task Offloading Model \cite{M.Hop.Tasks}, where they used a Bat Algorithm approach to determine the best candidates for task offloading of \acp{SWC}.
The aforementioned \ac{CAM} and \ac{BSM} are enhancing safety by sharing information like vehicle motion data (e.g. position) and state data. However, there are several approaches with the aim to extend the V2X application even further. 
Through the integration of external data provided by other edge devices, several functionalities can be executed in a coordinated manner through cooperative services, leading to enhanced safety in certain maneuvers like overtaking scenarios and intersection crossings \cite{Hobert2015}.

\subsection{Current state \ac{RSU}}\label{subsec:rsu}
In the development of \ac{ITS}, a \ac{RSU} is an integral component of the communication network. As they can be placed carefully at places needed, they support a more efficient exchange of information between vehicles and smart infrastructure \cite{Maglogiannis2022}.
Current research mostly focuses on optimizing the deployment of \ac{RSU}, e.g., how they can be evenly distributed in an area while still be cost-efficient. 
A graph based approximation algorithm which includes different RSU types based on static as well as mobile \ac{RSU} attached to public transportation and specifically controlled vehicles to assist was introduced for \ac{VANET} systems.  

%% file: sections/03_Concept.tex
\section{Concept}\label{sec:concept}

\begin{figure}
    \centering
    \includegraphics[width=1\linewidth]{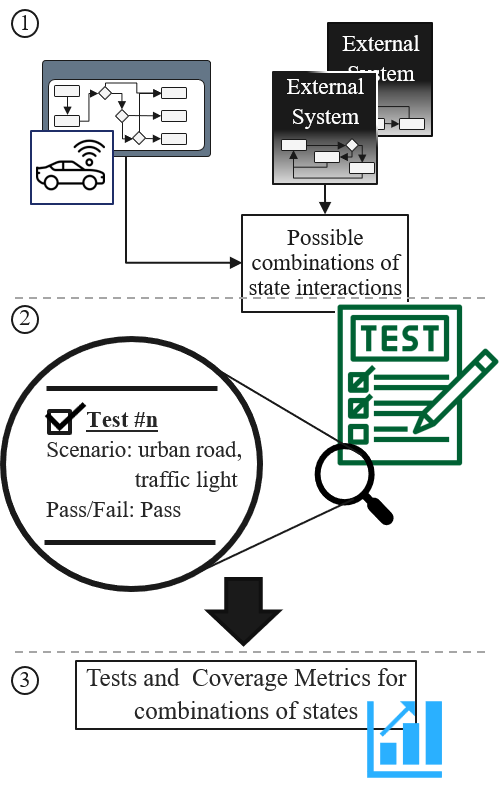}
    \caption{Concept of the state chart method}
    \label{fig:concept}
\end{figure}

The development of driving functions that incorporate different subsystems, such as external sensors, presents a challenge due to the increase in the combinatorics of different states of interaction between those systems, also known as the \textit{curse of dimensionality}. In addition to an increase in dimensionality, the information from systems outside the ego vehicle can be unreliable, incomplete, or unknown. Connected systems for \ac{AD} can involve \acp{RSU} and traffic lights, for which complete system information may not be available. The necessary \ac{VV} of such networked systems represents a bottleneck. Modeling the flow of information from external systems and focusing on the possible states of the overall driving function in these interactions mitigates some of the challenges posed by the dimensionality of the system. The proposed method consists of three distinct steps, shown as circled numbers in \cref{fig:concept}. 

\textbf{Step 1 -} All involved systems, internal and external to the vehicle and relevant to the driving function under development, are identified. Only the interaction between the vehicle and these subsystems is specified and modeled.
This allows developers to model external systems as a gray box, see \cref{fig:concept}, where the interaction with the vehicle system is known and can be defined, but the inner workings of each external system can be simplified or omitted. Focusing on the flow of information and not the individual messages between subsystems enables developers to incorporate different external systems. 
State charts offer a way to model the interaction in such systems. The flow of information and the different states of interaction for each subsystem can be defined by excluding states that are not directly involved in the interaction between an external system and the ego vehicle. As transitions between states in a state chart are based on events and the previous state, it is possible to place greater emphasis on the flow of information, rather than the overall system itself. Through this modeling, all possible interactions between subsystems and their corresponding states in each state chart can be identified. This enables the definition of the scenario space of potential driving scenarios. Verifying and validating such systems requires assessing the interactions between subsystems and their impact on the driving function as a whole. 

\textbf{Step 2 -} By linking the established scenario-based testing with the interaction between subsystems, modeled as state charts, the \ac{VV} effort can be reduced due to the simplification of external systems and the focus on the effect of different information on the ego vehicle. Even with these reduced subsystem models, the overall number of possible state combinations is increasing rapidly.
State charts are capable of being subjected to unit tests, as shown in \cref{fig:concept}, thereby allowing for their verification. Scenarios and the involved states for each subsystem can be linked to those unit tests. This enables tests that focus only on the involved states for a given scenario. Conversely, it is possible to identify the states of the individual subsystems in observed scenarios and link them with corresponding unit tests.

\textbf{Step 3 -}
This allows the space of possible state combinations to be tested and statements to be made whether enough of these scenarios have been observed to pass this test. 
This approach can be extended with new edge devices and other services, as only the interaction between the new edge system and the existing state chart must be defined and modeled.

The structuring of the interaction in connected systems with state charts and the linking of those state charts to scenarios enables the straightforward definition of driving function tests. The specification of which states are involved in a given scenario allows the coverage of test cases and scenarios to be determined. By focusing on the flow of information and the interaction between subsystems, new systems can be modeled as gray boxes and incorporated with relative ease.
This work demonstrates the proposed methodology using an inner-city crossing equipped with a \ac{RSU} which is capable of detecting and locating other \acp{VRU} and a traffic light broadcasting its current state. Two exemplary use cases are analyzed in this work. 
In use case 1, observed driving scenarios are matched to possible combinations of different states of the model, allowing for the identification of areas with inadequate scenario coverage. 
In use case 2, the testability of state charts is demonstrated using unit tests.
For this use case, the occurrence of jaywalking pedestrians and the RSU-based detection and localization of these pedestrians are formulated as tests and evaluated using the recorded scenarios.


%% file: sections/04_Implementation.tex
\section{Implementation}\label{sec:implementation}

\begin{figure}[h]
    \centering
    \includegraphics[width=1\linewidth]{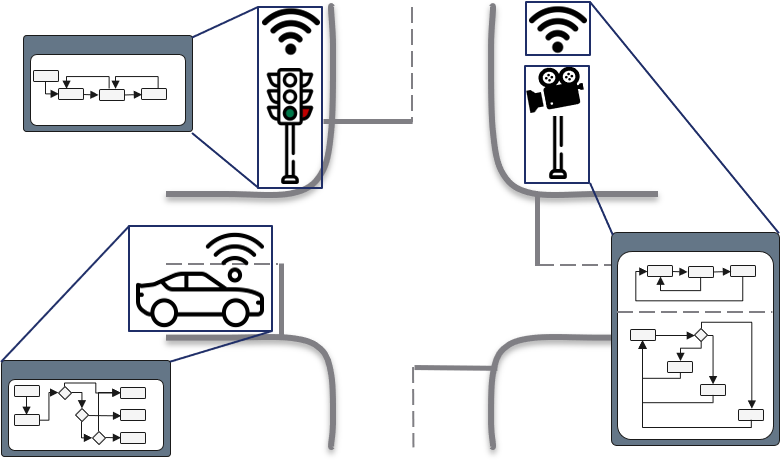}
    \caption{Connected Systems as state charts}
    \label{fig:story_setup}
\end{figure}

The objective of this implementation is to demonstrate the testability and reduction in dimensionality for connected systems using the proposed state chart-based method. To this end, an exemplary traffic situation at an intersection is examined. An automated vehicle requests information from external sensors in the form of a \ac{RSU} detecting \acp{VRU}. Additional information about the traffic situation is provided by a traffic light broadcasting its current signal state.
The three simplified subsystems (vehicle, \ac{VRU}, traffic light) are simulated as interacting and exchanging information. 
A structured approach using scenario mapping and unit tests is employed to manage this dimensionality. 
Scenario mapping can provide information about the coverage of system behavior. Unit testing helps verify that each component works correctly in isolation. 
The applicability of the proposed approach will be evaluated using scenarios generated in the CARLA simulation platform. 
The scenarios take place at a signalized four-way junction and encompass the trajectories of all participants, including vehicles and pedestrians. 
This approach allows for a detailed analysis of how different elements interact in a controlled traffic environment. \Cref{fig:story_setup} shows the involved systems.

\subsection{Scenario generation}\label{subsec:scenario generation}

Scenarios for this implementation were generated using CARLA 0.9.15. A two-lane, four-way junction equipped with pedestrian crossings and traffic signals was chosen as the location of the scenario generation. 
CARLAs inbuilt traffic generation was used to spawn vehicles of various types and pedestrians. Among the pedestrians, some were designated as jaywalkers, introducing unpredictable elements into the simulation (see \Cref{tab:simulation_hyperparameters}).
Each scenario describes the trajectory of an object (vehicle or pedestrian) within the intersection. All participants active at a given time are included in a scenario, providing a comprehensive view of the interactions occurring for the whole intersection. Of these, 9,484 scenarios specifically involve cars navigating through the intersection. The specific routes taken by these cars were identified. They serve as ego vehicles and are the main actors in each scenario. The \ac{RSU} information and traffic signal states are emulated with regard to these 9,484 ego vehicles.

\begin{table}[h]
    \caption{Hyperparameters for scenario generation}
    \label{tab:simulation_hyperparameters}
    \begin{tabular}{ll}
    \toprule
        Hyperparameter & Value \\
        \midrule
        CARLA Version & v0.9.15\\
        Number of vehicles & 80\\
        Number of pedestrians & 50\\
        Percentage of jaywalking pedestrians & 10\%\\
        Number of generated Scenarios & 27,709\\
        \midrule
        Number of ego-scenarios & 9,484\\
    \bottomrule
\end{tabular}
\end{table}

\subsection{Roadside unit and traffic light modeling}\label{subsec:rsu_and_lights}

Both systems serve to add dimensionality to the simulation, and therefore are emulated in a highly simplified manner using the following assumptions. It is assumed that the \ac{RSU} can see and potentially detect all \acp{VRU} at the crossing. The \ac{RSU} is assumed to either detect a \acp{VRU} presence or locate it in the next step. The detection process is emulated with a 90\% success rate, while the localization of the exact position is emulated with a 75\% success rate after detecting the \acp{VRU}. These values have been chosen intentionally low in order to reduce the necessary number of scenarios needed to evaluate the approach. \Cref{fig:statechart} in the appendix illustrates the detailed state chart of the localization.
Furthermore, traffic signals are emulated as continuously broadcasting their state to all participants. Signals relevant to the ego vehicle are extracted, and state changes (e.g., from red to green) are considered. This simplification facilitates the evaluation process while still providing meaningful insights into performance of a system.

\subsection{State charts for each subsystem}\label{subsec:state charts}
The interaction between the \ac{RSU}, the ego vehicle and the traffic lights are modeled as state charts. \Cref{fig:statechart} shows the individual state charts where each individual subsystem is described in the text below. 

\textbf{Vehicle state chart -}
The vehicle approaches the crossing and communicates with the \ac{RSU} requesting information about all participants at the crossing. 
Information about the state of the traffic light is constantly received. 
Depending on the return signal and the success of the transmission between \ac{RSU} and the vehicle, different driving profiles, e.g., free turn, caution or stop, are selected. 

\textbf{\ac{RSU} state chart -}
The functionality of the \ac{RSU} is divided into two parts: \textit{localization} and \textit{\ac{V2I} communication}. 
The behavior of the \ac{RSU} is modeled as follows: when a \ac{VRU} enters the area of the \ac{RSU}, the \ac{RSU} detects the \ac{VRU} with an unknown percentage of accuracy. 
The \ac{RSU} can then either locate the position of the \ac{VRU} or fail to do so. 
When the \ac{RSU} is requested for information, it sends the current list of detected and/or localized \acp{VRU} to the approaching vehicle. 
The transmission can be successful or unsuccessful.

\textbf{Traffic light state chart -}
The traffic light is modeled as a simple transition between each signal phase and can be easily specified in more detail by adding the times of each individual phase.
In this example, it is assumed that the signal state is constantly broadcasted and received by the vehicle. 

\Cref{tab:states} shows the number of possible states in each subsystem and the total number of possible state combinations for the whole system. 
The \textit{RSU Communication} can transmit each state either successfully or unsuccessfully, resulting in seven possible outcomes of this communication. Scenarios span multiple seconds, leading to multiple states of the traffic light occurring during a scenario. Therefore, the transitions between states of the traffic light are considered as additional, combined states of the state chart. They are red to yellow, yellow to red, yellow to green and green to yellow.

\begin{table}[h]
  \caption{Combination of possible states}
  \label{tab:states}
  \begin{tabular}{ll}
    \toprule
    Subsystem & number of possible states\\
    \midrule
    Traffic light & 4 + 4\\
    \ac{RSU} localization & 3\\
    \ac{RSU} communication & $3 \cdot 2 + 1$\\
    vehicle & 5\\
    \midrule
     overall number of states& 840\\
  \bottomrule
\end{tabular}
\end{table}    

%% file: sections/05_Evaluation.tex
\section{Evaluation}\label{sec: evaluation}
\subsection{Use case 1: Coverage analysis}\label{subsec:coverage}

\begin{figure*}[htb]
    \centering
    \includegraphics[width=0.94\textwidth]{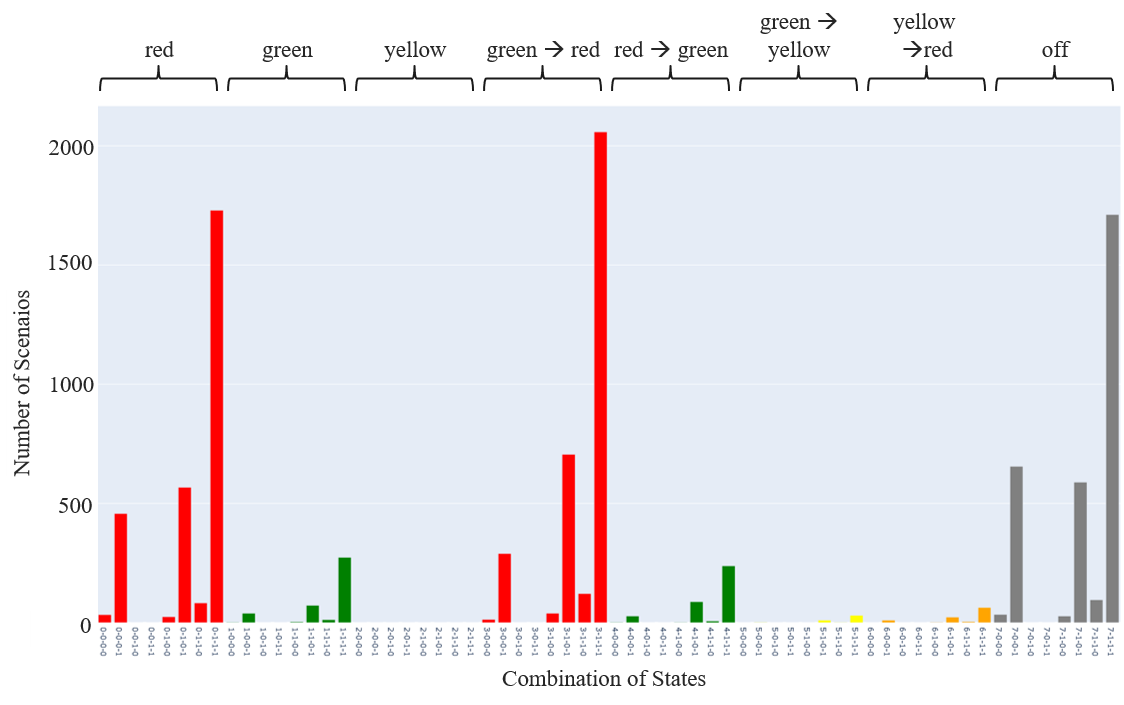}
    \caption{Coverage of possible state combinations}
    \label{fig:coverage}
\end{figure*}

Use case 1 describes a situation in which observations are made at an intersection (here, simulated scenarios) to determine whether scenarios covering all relevant combinations of states of the connected driving function are observed. To achieve this, the states of the traffic light and the transitions between the traffic light phases were considered for all ego scenarios. \Cref{fig:coverage} shows the number of scenarios that match the respective combinations. The color of the bars indicate the state of the traffic light, as well as the first number in the numerical code below the bars. Additionally, the states \textit{\ac{VRU} detected}, \textit{\ac{VRU} located} and \textit{transmission successful}, see \Cref{fig:statechart}, are noted in boolean notation in the numerical code below the bars. In this reduced space of possibilities, there are 64 combinations. 
\Cref{fig:coverage} clearly shows that the observed scenarios are not equally distributed across all combinations of states. Rather than observing scenarios until all possible combinations of states are sufficiently covered, it is necessary to inquire into specific combinations of states. As an example, it would be beneficial to determine the circumstances under which uncertain information prompts the driving function to transition to driving profile 1 in the state chart. The objective is to identify the instances when and how often these situations occur and the system states involved. For example, when the states are \textit{traffic light = red}, \textit{VRU detected = True}, \textit{VRU located = False}, and \textit{transmission successful = True} (numerical code 0-1-0-1 in figure \ref{fig:coverage}), it leads to \textit{Driving Profile 1}. This scenario, in which information about present but not exactly located \acp{VRU} is transmitted successfully, occurs 567 times. Conversely, the issue where a \ac{VRU} is detected but the transmission fails is rare, occurring only 24 times in the observed scenarios (numerical code 0-1-0-0). These combinations of \ac{VRU} states do not occur at all while the traffic light is yellow and only occur in small numbers while the traffic light is transitioning from or to a yellow signal. 
Although it is possible to make coverage statements (e.g., more scenarios from yellow traffic light phases are necessary), it becomes evident that a more targeted selection of scenarios and state combinations is necessary. 

\subsection{Use case 2: Test case generation}\label{subsec:unit_tests}

\begin{figure*}[hbt]
    \centering
    \includegraphics[width=0.94\linewidth]{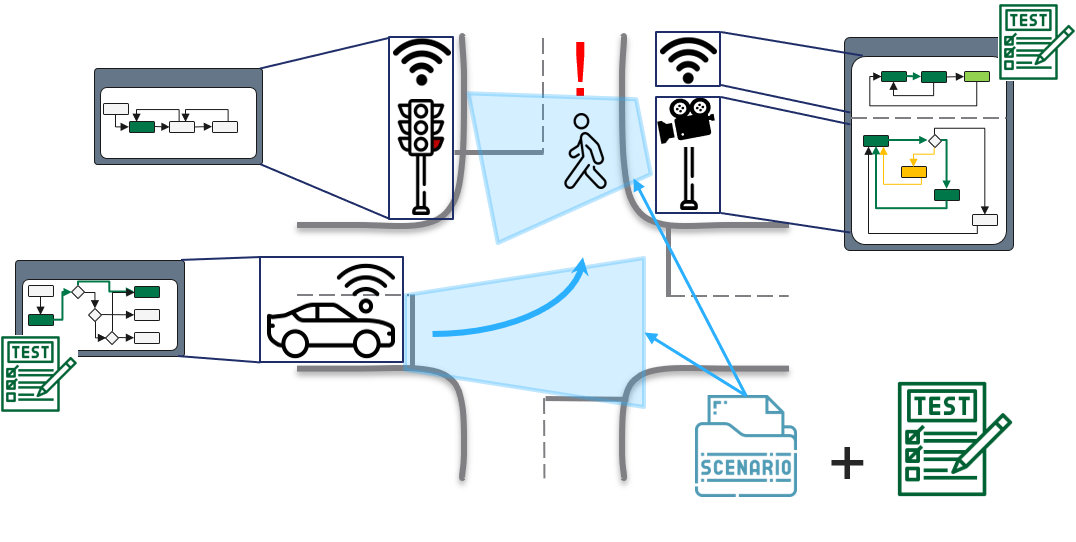}
    \caption{Unit tests for specific scenarios}
    \label{fig:story_payoff}
\end{figure*}

Testing every possible combination of states in a connected, automated system, as shown in \cref{fig:coverage}, is not feasible. Even with simplified subsystems the possible combination of states ranges in the hundreds as shown in \cref{tab:states}. For the interaction modeled in a set of state charts unit tests can be defined. \Cref{fig:story_payoff} shows a unit test suite and the corresponding scenario.

For use case 2 the combination of states that lead to unclear information (\textit{Possible \ac{VRU} present}, \textit{driving profile 1}) in \cref{fig:unit_tests} is examined. Observed scenarios can then be assigned to each of these tests. State chart tools (Itemis CREATE was employed in this work) were utilized to identify which states are included within the scope of the defined tests. This process enables the determination of the coverage of interactions defined in unit tests between these subsystems. By allocating scenarios to these tests, statements can be made about the coverage of the scenarios within the individual tests.
First, a set of four unit tests (see \cref{fig:unit_tests}) with state combinations leading to the vehicle entering \textit{driving profile 1} is defined. All combinations of states leading to the vehicle state \textit{Possible VRU present} are covered by this set. The states of the traffic light do not directly lead to this state and are not considered for this use case. Through these tests the number of relevant states of the state charts for this use case is reduced. In 2511 of 9484 scenarios selected for the use case the interaction with the \ac{RSU} leads to unclear information about possible \acp{VRU} at the crossing. Unit tests 2, 3, and 4 represent situations where the information from the \ac{RSU} is not conclusive. These scenarios were observed less frequently than normal \ac{RSU} behavior. If more scenarios are required for testing, the scenario generation or observation can be guided by the scenario coverage for each test.
The traffic light and the information it provides do not directly impact the decision of the ego vehicle to enter different driving profiles. For unit tests 4.1 and 4.2, the phases of the traffic light were also considered. A total of 321 scenarios were assigned to unit test 4. Of these, only six occurred while the traffic light switched from green to red (for the ego vehicle), while 121 scenarios occurred when the traffic light switched from red to green.
The incorporation of data from subsystems, such as traffic lights, enables the refinement of unit tests and facilitates the targeted testing of functions and the targeted recording of scenarios.

%% file: sections/06_Conclusion.tex
\section{Conclusion}\label{sec:conclusion} 

This paper introduces a novel method for modeling interactions within connected, automated systems using state charts, with a particular focus on the flow of information between the ego-vehicle and external systems and sensors. The approach integrates state charts into established scenario-based testing methodologies. By combining state charts and their inherent testability with scenario-based testing, it can be demonstrated that the curse of dimensionality in connected systems can be managed more effectively. The methodology was explored through two distinct use cases, demonstrating the capability of the method for statements on scenario coverage. Use case 1 demonstrated the underlying issue of scenario coverage for all possible interactions between connected subsystems. The second use case demonstrated the application of unit tests for state charts for interactions and the mapping of scenarios to these tests. This illustrated how the method supports a more targeted validation and verification process. Additionally, it directs data collection efforts within connected systems to areas with inadequate scenario coverage. Future work will involve applying this method to real-world data rather than relying on simulated probabilities for \acp{RSU}, with the aim of further validating and refining the approach in practical environments.

%% file: sections/07_Appendix.tex
\section{Appendix}\label{sec:appendix}

\noindent\begin{minipage}{\textwidth}
    \centering    
    \includegraphics[trim={0 1.2cm 0 0}, clip, width=0.85\linewidth]{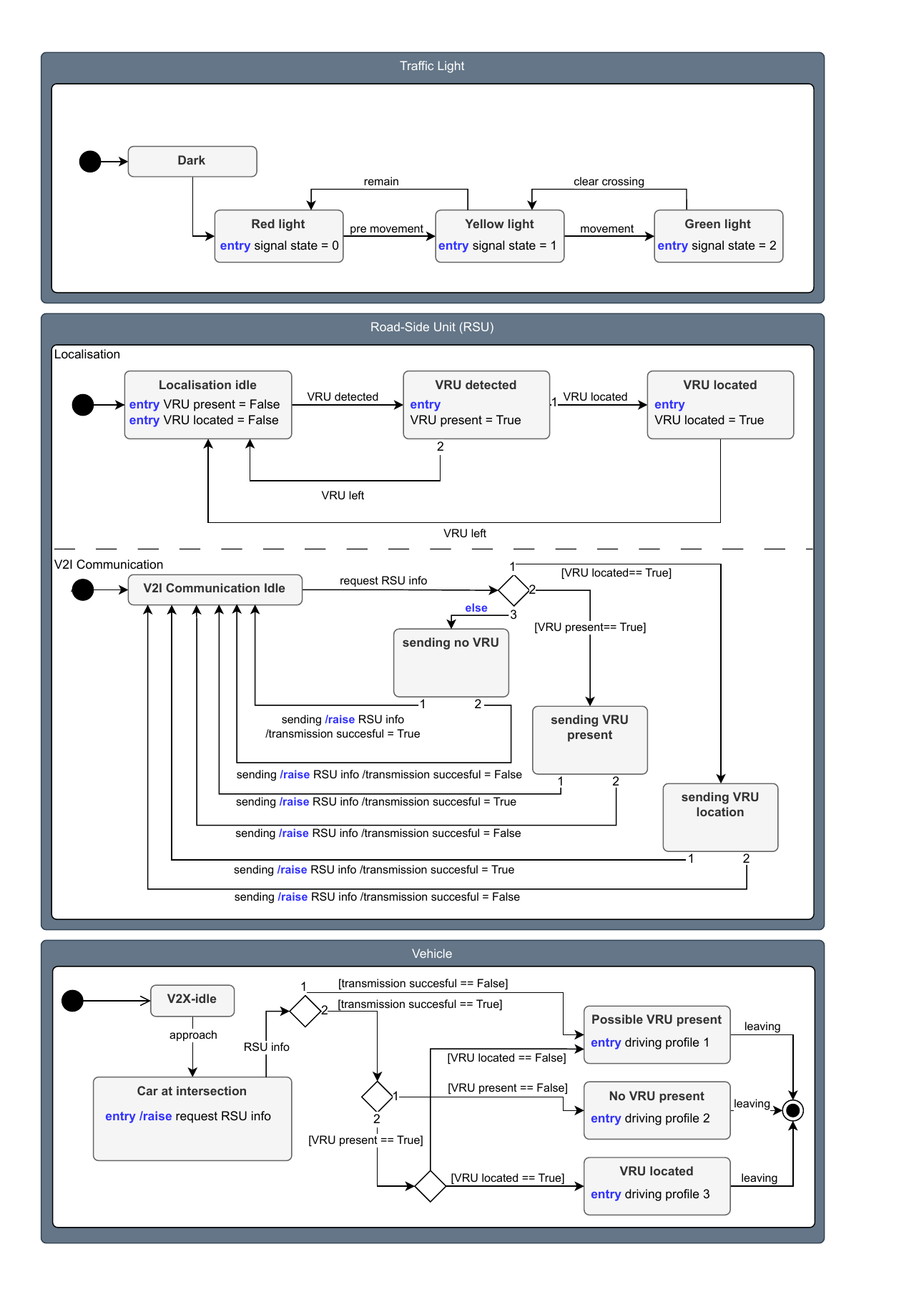}
    \captionof{figure}{Statecharts of each subsystem}    
    \label{fig:statechart}
\end{minipage}

\clearpage

\begin{figure*}[h!t]
    \centering
    \includegraphics[trim={0 1.7cm 0 1.5cm}, clip, width=0.93\linewidth]{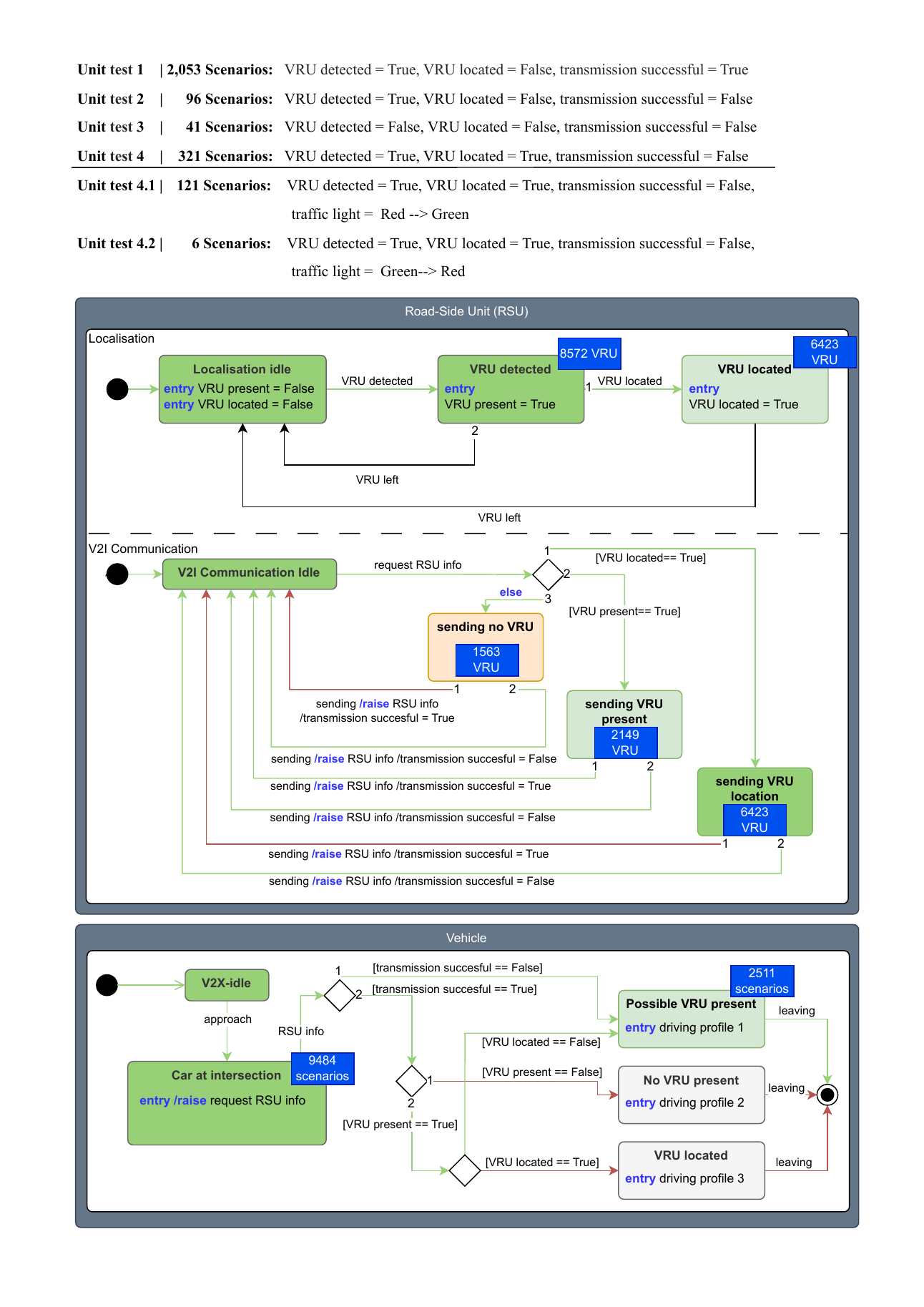}
    \caption{Unit tests for use case 2}
    \label{fig:unit_tests}
\end{figure*}